\pdfoutput=1

\documentclass[11pt]{article}

\usepackage{ACL2023}

\usepackage{times}
\usepackage{latexsym}

\usepackage[T1]{fontenc}

\usepackage[utf8]{inputenc}

\usepackage{microtype}

\usepackage{inconsolata}

\usepackage{graphicx}
\usepackage{multirow}
\usepackage{arydshln}

\newcommand{\mean}[1]{%
\ifdim#1pt<-10pt%
  \relax%
\else%
  \ifdim#1pt>0pt%
    \hphantom{-}%
    \ifdim#1pt<10pt \hphantom{1}\else\fi%
  \else%
    \hphantom{1}%
  \fi%
\fi#1}

\newcommand{\contrib}[1]{\ifdim#1pt<0.01pt \hphantom{0}\else\fi#1}

\title{
Exploring Anisotropy and Outliers in Multilingual Language Models for Cross-Lingual Semantic Sentence Similarity
}

\author{Katharina Hämmerl$^{1,2}$ \and
  Alina Fastowski$^{1}$  \AND
  Jindřich Libovický$^{3}$  \and
  Alexander Fraser$^{1,2}$ \\
  \\
$^1$Center for Information and Language Processing, LMU Munich, Germany \\
\texttt{\{haemmerl,fraser\}@cis.lmu.de}\\
$^2$Munich Centre for Machine Learning (MCML), Germany \\
$^3$Faculty of Mathematics and Physics, Charles University, Czech Republic \\
}

\begin{document}
\maketitle

\begin{abstract}

Previous work has shown
that the representations output by contextual language models are more
anisotropic than static type embeddings, and typically display outlier dimensions.
This seems to be true for both monolingual and multilingual models, although much less work has been done on the multilingual context.
Why these outliers occur and how they affect the representations is still an active area of research.
We investigate outlier dimensions and their relationship to anisotropy in multiple pre-trained multilingual language models.
We focus on cross-lingual semantic similarity tasks, as these are natural tasks for evaluating multilingual representations.
Specifically, we examine sentence representations.
Sentence transformers which are fine-tuned on parallel resources (that are not always available) perform better on this task, and we show that their representations are more isotropic. 
However, we aim to improve multilingual representations in general. 
We investigate how much of the performance difference can be made up
by only transforming the embedding space
without fine-tuning,
and visualise the resulting spaces.
We test different operations: Removing individual outlier dimensions, cluster-based isotropy enhancement, and ZCA whitening.
We publish our code for reproducibility.\footnote{\url{https://github.com/kathyhaem/outliers}}

\end{abstract}

\section{Introduction}

\begin{figure}
    \centering
    \includegraphics[width=.48\textwidth]{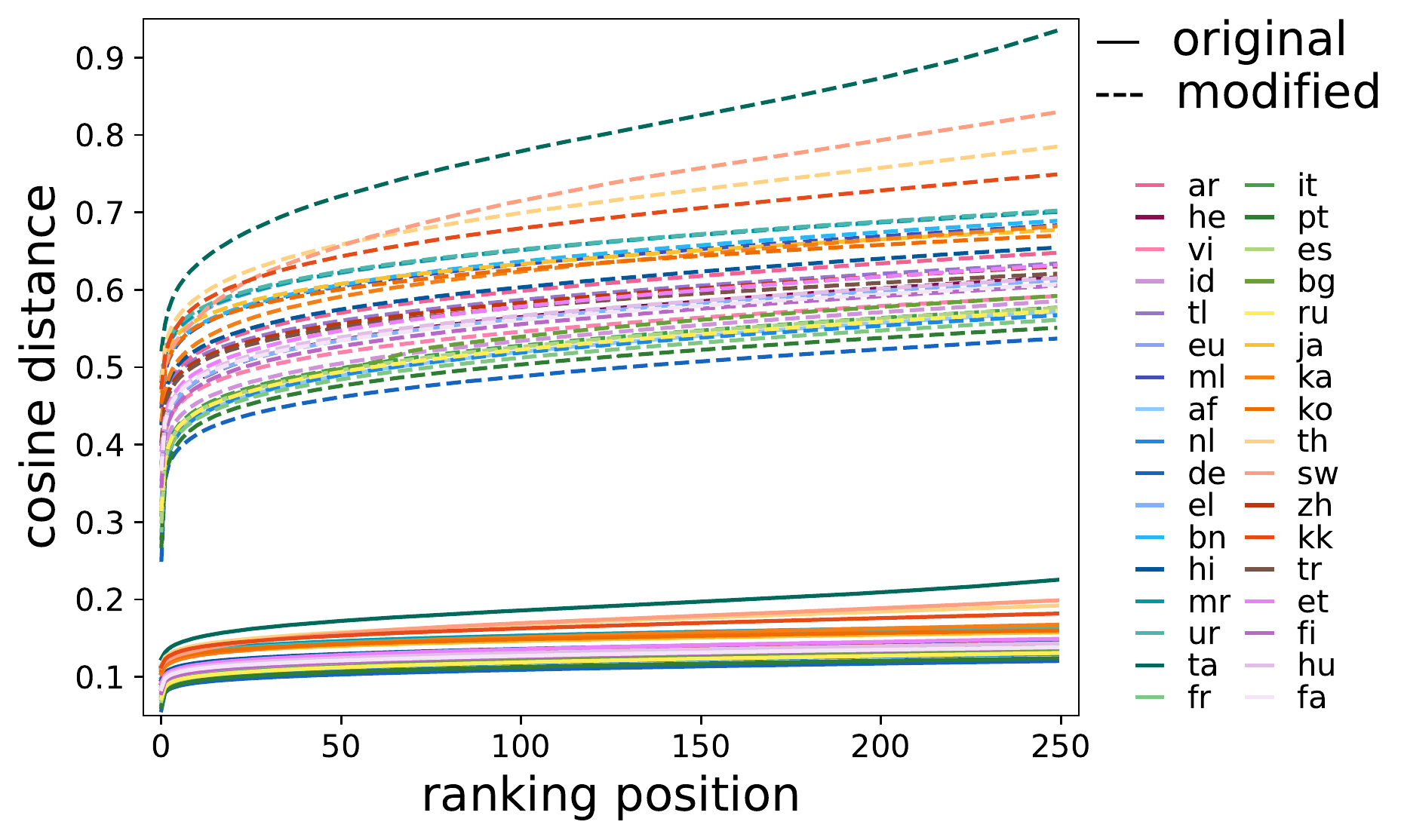}
    \caption{Effect of removing dimension 588 from layer 8 on Tatoeba cosine similarities. The x-axis is the ranking position of candidate sentences; the y-axis is their cosine distance from the query sentence, which should be relatively large for all but the correct translation.
    After removing 588 from the sentence representations, the highest ranking candidate sentence is much more clearly differentiated from the lower-ranking candidates.
    }
    \label{fig:cosines-588-306}
\end{figure}

Since BERT-like \citep{devlin-etal-2019-bert} language models rose to popularity,
much has been made of the study of their hidden states and parameters \citep[cf.][]{rogers-etal-2020-primer}.
Thanks to their ability to incorporate context, they have been a major improvement for most tasks over static input embeddings.
However, a certain issue has been shown in a number of works to affect contextual language models to a greater degree: outlier dimensions in the weights and hidden states \citep{kovaleva-etal-2021-bert} and correspondingly, high anisotropy (\citealp{Gao2019RepresentationDP, ethayarajh-2019-contextual}, inter alia).
At the same time, the raw pre-trained embeddings work surprisingly badly for semantic similarity tasks, prompting efforts to train better sentence embeddings such as 
done by
\citet{reimers-gurevych-2019-sentence}.

In this paper, we are interested in multilingual sentence embedding quality.
We discuss both outliers and anisotropy as two related aspects of embedding quality.
Outlier dimensions are typically defined as dimensions that consistently produce values of a magnitude more than three or five times the standard deviation of all dimensions \citep{kovaleva-etal-2021-bert}.
If a model has outlier dimensions in its hidden states, it will necessarily have higher anisotropy, since these dimensions create a consistent shift towards a certain direction in the embedding space.
On the other hand, high anisotropy can also occur without individual dimensions meeting the outlier definition, namely if some principal components composed of multiple dimensions are much larger than others.
Therefore, as we understand it, anisotropy is the wider phenomenon of which outliers are a subset.

From a theoretical perspective, high anisotropy is considered a problem because it means that the model is not using the full representation space available, and because it translates to high average cosine similarity even between unrelated words or sentences.
Figure~\ref{fig:cosines-588-306} illustrates this problem clearly.
This can increase the odds of picking a wrong candidate on word and sentence similarity tests, and makes representations produced by the model less expressive and less interpretable.

Outlier dimensions, since they contribute to anisotropy, entail similar challenges.
On the other hand, they are easy to spot, easy to manipulate, and a straightforward entry point to the anisotropy issue.
Previous work has sometimes found that models rely strongly on outlier weights for certain tasks, and are overly vulnerable to pruning a select few weights, e.g. \citep{kovaleva-etal-2021-bert}. 
Further, outliers have been found to present a challenge in model quantisation \citep{bondarenko-etal-2021-understanding}.

Because they are aspects of the output representations, studies of anisotropy and outliers often use semantic similarity tests that rely directly on these representations, without fine-tuning the model.
We follow this approach as well.
In this work, we specifically consider sentence representations.

Only a small amount of work has been done on outliers and isotropy in multilingual models, which we focus on.
\citet{rajaee-pilehvar-2022-isotropy} found that mBERT does not contain outlier dimensions, while XLM-R does.
However, both models nevertheless exhibit high anisotropy. 

Another important aspect to consider in the multilingual case is that even if representations are more or less isotropically distributed, the subspaces for different languages can still be misaligned, which further affects cross-lingual performance.
Training with parallel data, as done in \citet{reimers-gurevych-2020-making}, is one way to radically improve cross-lingual alignment.
However, we are interested in pushing models to perform well without parallel data.
The present work therefore
attempts to separate the effect of anisotropy from other factors that could account for the performance gap, such as the use of parallel data objectives and internal misalignment of languages.

\paragraph{Our contributions.}
This work provides an in-depth exploration of outlier dimensions and anisotropy in XLM-R and other pre-trained multilingual language models, using the Tatoeba \citep{artetxe-schwenk-2019-massively}, multilingual STS \citep{cer-etal-2017-semeval},
and BUCC 2018 \citep{ZWEIGENBAUM18.12}
semantic similarity tasks
and looking directly at the relevant hidden state representations. 

We confirm that certain outlier dimensions have a negative effect on similarity search in the cross-lingual setting (\S~\ref{sec:zeroing-dims}).
We find that outlier dimensions can differ between languages, although the largest outliers occur in all or most tested languages (\S~\ref{sec:zeroing-dims}). 
Anisotropy also varies across languages, and we observe a possible relationship to pre-training data size (\S~\ref{subsec:outlier-analysis}).
In our experiments, mBERT does exhibit outlier dimensions (\S~\ref{subsec:outlier-analysis}).

Looking at semantic similarity task performance, we show that zeroing outliers and isotropy-enhancing transformations are quick ways to improve model performance on such tasks (\S~\ref{sec:zeroing-dims},~\ref{subsec:isotropy-enhancing}).
However, a multilingual sentence-transformer performs much better out-of-the-box, and benefits little to not at all from further increasing isotropy.
As we show in \S~\ref{subsec:outlier-analysis}, this model is already much more isotropic than XLM-R, its pre-trained equivalent.

Finally, we give a clearer intuition of the phenomena in question by using tSNE \citep{Maaten2008VisualizingDU} to visualise embedding spaces (\S~\ref{subsec:tsne}).
This allows us to grasp more intuitively how anisotropy is one aspect of misalignment between languages in multilingual models.

\section{Related Work}

BERT-like models have dominated NLP research in recent years.
Multilingual BERT \citep{devlin-etal-2019-bert} and XLM-R \citep{conneau-etal-2020-unsupervised} are two popular models whose variants are used for many different ends.
Accordingly, some amount of research has focused on analysing properties of the models, sometimes called ``BERTology'' \citep{rogers-etal-2020-primer}.
The phenomena we discuss in this paper---outlier dimensions and anisotropy---are just two aspects of model analysis.

\subsection{Describing the phenomena}

First, we discuss outlier dimensions specifically.
\citet{kovaleva-etal-2021-bert} focus specifically on outlier dimensions in the LayerNorm weights of English BERT.
Around the same time that the LayerNorm outliers arise, training loss and evaluation perplexity start to fall off sharply.
The exact cause is unknown but this suggests the outliers help the model, which they corroborate by showing that task performance decreases significantly when zeroing out outlier weights after fine-tuning.
If zeroing the weights is done before fine-tuning, the model recovers most of the performance, but a slight disadvantage is still observed.

\citet{timkey-van-schijndel-2021-bark} take a different view of outliers in that they analyse hidden representations instead of weights.
They also focus on similarity measures and find that in this context, the outlier dimensions ``obscure representational quality''.
\citet{rajaee-pilehvar-2022-isotropy} are one of few to focus on outliers in multilingual models:
They find no outliers in mBERT, but do find them in XLM-R.
The paper also looks at the embeddings of different languages separately, an approach we follow for the majority of our experiments.
%

As we mention above, outliers are one way to look at anisotropy in hidden representations.
\citet{ethayarajh-2019-contextual} is one of the first to present evidence for unusually high anisotropy in contextual embedding models, including BERT and GPT-2.
\citet{Gao2019RepresentationDP} describe the \textit{representation degeneration problem} and suggest using cosine regularisation to mitigate it.
We discuss mitigation approaches in more detail below (\S~\ref{subsec:mitigate}).

There are multiple ways to measure (an)isotropy, including but not limited to:
\begin{itemize}
    \item average cosine similarity \citep[cf.][]{ethayarajh-2019-contextual, timkey-van-schijndel-2021-bark}
    \item based on principal components \citep{mu-viswanath-2018-abtt}
    \item IsoScore \citep{rudman-etal-2022-isoscore}
\end{itemize}

These are continuous measures, with value ranges depending on the method.
While lower anisotropy is theoretically desirable, it can be hard to decide at what point a space is ``isotropic enough''.
In the present work, we stick to the first measure, that is, average cosine similarity between random pairs (see \S~\ref{subsec:outlier-analysis}).

\subsection{Searching for causes}

It has been shown that word frequency plays a significant role in how representations are distributed in contextual models: For instance, rare words tend to be pushed further from the origin during pre-training, leading to a separation of tokens by frequency.
\citet{yu-etal-2022-rare} show that rare token embeddings are the first to become anisotropic during pre-training, and seem to ``take down with them'' the rest of the space.
\citet{puccetti-etal-2022-outlier} similarly find that outliers are ``driven by token frequency''.

On the other hand, \citet{luo-etal-2021-positional} argue that outliers are caused by positional embeddings which display outliers, and this propagates forward through the model.
They demonstrate this by training RoBERTa models with and without positional embeddings.
The model without positional embeddings has much worse perplexity, but no outliers.
This idea has not been confirmed by other works, 
and \citet{rajaee-pilehvar-2022-isotropy} find that multilingual BERT, despite having positional embeddings, does not display outliers.
We use a different mBERT checkpoint in our experiments which does exhibit outliers, but we draw no conclusions about positional embeddings.

\subsection{Attempts at mitigation}\label{subsec:mitigate}

Various methods have been suggested to increase isotropy in the contextual embedding space.

\paragraph{During training.}
\citet{Gao2019RepresentationDP}, who described anisotropy early on, proposed a cosine regularisation term to mitigate it.
This term simply maximises the angle between any non-identical words.
Building on this, \citet{zhang-etal-2020-revisiting}
propose Laplacian regularisation as a way to specifically reduce similarity of word pairs that do not occur in similar contexts.
\citet{Ferner2022BenefitsFV} apply a token-level variational loss to an encoder-decoder Transformer, similar to what is done in Variational Auto-Encoders.
All three works add the regularisation terms to a model they train from scratch.

On the other hand, \citet{ding-etal-2022-isotropy} test several BERT-like models on GLUE tasks before and after ``isotropy calibration'' (fine-tuning with regularisation terms), and find that task scores do not consistently improve.
They reason that this is because the models already benefit from local isotropy, thus further isotropy calibration does not help.
We also note that these experiments are all done on tasks that use fine-tuning.

\paragraph{Post-hoc.}

Rather than training a model from scratch,
\citet{li-etal-2020-sentence} train normalising flows on STS and similar datasets that they want to test on, starting with a pre-trained BERT model--- they call this approach \textit{BERT-flow}. 
Both \citet{Su2021WhiteningSR} and \citet{huang-etal-2021-whiteningbert-easy} apply whitening to sentence representations.
This operation transforms the mean of the sentence vectors to zero, and the covariance matrix to the identity matrix, as we discuss in more detail in \S~\ref{subsec:isotropy-enhancing}.
\citet{Su2021WhiteningSR} combine this with a dimensionality reduction strategy.

\citet{timkey-van-schijndel-2021-bark} also test several ways of postprocessing representations, such as standardisation and removing the top few principal components.
\citet{liang-etal-2021-learning-to-remove} and \citet{rajaee-pilehvar-2021-cluster} remove dominant directions from the embedding space.
The former learns a set of parameters for weighted removal (scaling) of principal components, while the latter clusters the data before removing the top principal components from each cluster.
\citet{rajaee-pilehvar-2021-fine-tuning} find that removing the dominant directions after S-BERT training decreases STS performance, while removing them from the vanilla model improves performance. We corroborate these findings for the multilingual case.
\citet{Jung2022IsotropicRC} apply isotropy-improving methods, namely normalising flows and Whitening, in the context of dense retrieval models, and find score improvements on the target task.

\paragraph{Contrastive fine-tuning.} Contrastive learning has become a popular technique in NLP in recent years \citep{zhang-etal-2022-contrastive-data}.
Among other things, it has been shown to improve sentence embeddings and ensure they are more uniformly distributed.
Examples include \citet{gao-etal-2021-simcse, kim-etal-2021-self, zhang-etal-2021-pairwise, yan-etal-2021-consert}, and \citet{reimers-gurevych-2019-sentence}.
The latter, which we use as a reference in this work, uses in-batch contrastive optimisation in later implementations.

\section{Datasets}

Because we will show results of each of our experiments as we go along, we start here by introducing the datasets used.

\subsection{Tatoeba} 
This is a cross-lingual sentence retrieval task compiled by \citet{artetxe-schwenk-2019-massively} and pruned to 36 languages by \citet{hu2020xtreme}.
We follow the implementation used by the latter.
Each language is matched with English, and the objective is to find the correct translation for each query.
The subtasks per language contain 1k examples each.
The most similar translations are retrieved using the cosine similarity of the mean-pooled hidden representations from layer eight.
The metric is accuracy.

\subsection{BUCC} 
This is another similarity search task introduced by \citet{ZWEIGENBAUM18.12}.
However, since it focuses on parallel corpus building, not every query sentence has a match in the target language.
Therefore, both precision and recall are important to performance.
BUCC has four subtasks: German-English, French-English, Russian-English, and Chinese-English.
We again follow the implementation by \citet{hu2020xtreme}.
The test data contains several hundred thousand examples in each corpus, with between 1900 (Chinese) and 14400 (Russian) matched pairs.
The task metric is F1.

\subsection{Multilingual STS}
Another cross-lingual semantic similarity task is Multilingual STS \citep{cer-etal-2017-semeval} from SemEval 2017.
The task here is to score sentence pairs on a scale from 0 to 5
representing their relative similarity.
There are four cross-lingual subtasks, namely Arabic-English, two Spanish-English tasks of varying difficulty, and Turkish-English.
Each subtask contains 250 examples.
The task metric is Pearson correlation with the gold labels.

\subsection{Wikipedia}
Following \citet{rajaee-pilehvar-2022-isotropy}, we further use a sample of Wikipedia data in six languages (Arabic, English, Spanish, Sundanese, Swahili, and Turkish) for our analysis. We use these for comparability, as we investigate some of the same multilingual models.
The datasets contain between 347 (Sundanese) and 4952 (English) sentences.

\section{Outlier and anisotropy analysis}\label{subsec:outlier-analysis}

\begin{table*}[th]
    \centering
    \begin{tabular}{c|c|c|c|c}
        \textbf{Model} & \textbf{Anisotropy} & \textbf{Outliers} & \textbf{Means} & \textbf{Mean Cosine Contribution} \\
        \hline
        \multirow{4}{4em}{\mbox{XLM-R}} & \multirow{4}{1.5em}{0.92} & 588 & \mean{-15.18} & \contrib{0.77} \\
        & & 306 & \mean{3.08} & \contrib{0.03} \\
        & & \textit{239} & \textit{\mean{-2.06}} & \textit{\contrib{0.02}} \\
        & & \textit{180} & \textit{\mean{1.86}} & \textit{\contrib{0.01}} \\
        \hline
        \multirow{3}{4em}{mBERT} & \multirow{3}{1.5em}{0.73} & 227 & \mean{-11.64} & \contrib{0.39} \\
        & & 195 & \mean{-8.01} & \contrib{0.16} \\
        & & \textit{731} & \textit{\mean{2.70}} & \textit{\contrib{0.02}} \\
        \hline
        \multirow{5}{4em}{Multil. \mbox{S-BERT}} & \multirow{5}{1.5em}{0.35} & 588 & \mean{-6.78} & \contrib{0.22} \\
        & & 145 & \mean{-1.54} & \contrib{0.02} \\
        & & \textit{306} & \textit{\mean{1.46}} & \textit{\contrib{0.003}} \\
        & & \textit{459} & \textit{\mean{-1.43}} & \textit{\contrib{0.01}} \\
        & & \textit{741} & \textit{\mean{1.21}} & \textit{\contrib{0.01}} \\
    \end{tabular}
    \caption{Outliers and anisotropy scores in layer 8 of each model. The numbers in this table are based on Tatoeba data. Outliers are sorted by magnitude. 
    We show all outliers according to the $3\sigma$ definition of outlier dimensions. We \textit{italicise} dimensions that do not qualify as outliers under the $5\sigma$ definition.
    }
    \label{tab:outlier-anisotropy-1}
\end{table*}

Starting with data from Tatoeba,
we derive sentence embeddings for all statements in each dataset.
By \emph{deriving sentence embeddings}, we mean encoding each sentence using the model's standard tokeniser, running it through the model in inference mode, then mean-pooling the result while
ignoring special tokens. 
We proceed to calculate anisotropy scores for each language and dataset, as well as the outlier dimensions.
We use the $3\sigma$ definition of outliers here.
Note, however, that by considering sentence embeddings, which are already mean-pooled in one direction, we essentially have a smaller standard deviation and thus a more sensitive measure.
For this reason, we also show which outliers are smaller than $5\sigma$ by \textit{italicising} them in our tables. 

For the anisotropy score, we adapt \citeposs{timkey-van-schijndel-2021-bark} definition to the sentence level.
Let $S$ be a sample of $n$ random sentence pairs from a corpus $D$.
The approximate anisotropy $A(f_l)$
of layer $l$ in model $f$ is then:
\begin{equation}
    A(f_l) = \frac{1}{n} \cdot \sum_{\{x, y\} \in S} \cos(f_l(x), f_l(y))
\end{equation}
where $\cos(u, v)$ is the cosine similarity.

Further, we calculate the contributions to anisotropy of the largest dimensions.
Analogously to the overall anisotropy, if $CC_i(u, v) = \frac{u_i v_i}{\|u\| \ \|v\|}$ is the contribution of dimension $i$ to the total cosine similarity of $u$ and $v$,
then the contribution of dimension $i$ to the overall anisotropy is:

\begin{equation}
    CC(f^i_l) = \frac{1}{n} \cdot \sum_{\{x, y\} \in S} CC_i(f_l(x), f_l(y)).
\end{equation}

We use hidden representations from layer 8 when applying these techniques on Tatoeba data,
since this task is usually done using layer 8.
We test XLM-R, mBERT, and a multilingual S-BERT \citep{reimers-gurevych-2020-making} model which we have found to create good sentence embeddings across many languages.\footnote{The specific model we used is \texttt{sentence-transformers/xlm-r-100langs-bert-base\-nli-stsb-mean-tokens} and can be found on Huggingface.}

\begin{table*}[th]
    \centering
    \begin{tabular}{c|c|c|c|c}
        \textbf{Model} & \textbf{Anisotropy} & \textbf{Outliers} & \textbf{Means} & \textbf{Mean Cosine Contribution} \\
        \hline
        \multirow{2}{4em}{\mbox{XLM-R}} & \multirow{2}{1.5em}{0.99} & 588 & \mean{17.86} & \contrib{0.89} \\
        & & 741 & \mean{-5.62} & \contrib{0.09} \\
        \hline
        \multirow{7}{4em}{mBERT (cased)} & \multirow{7}{1.5em}{0.61} & 423 & \mean{-1.97} & \contrib{0.03} \\
        & & 731 & \mean{-1.54} & \contrib{0.02} \\
        & & \textit{373} & \textit{\mean{-1.22}} & \textit{\contrib{0.01}} \\
        & & \textit{89} & \textit{\mean{-1.04}} & \textit{\contrib{0.01}} \\
        & & \textit{511} & \textit{\mean{-0.99}} & \textit{\contrib{0.01}} \\
        & & \textit{761} & \textit{\mean{-0.92}} & \textit{\contrib{0.01}} \\
        & & \textit{493} & \textit{\mean{-0.86}} & \textit{\contrib{0.01}} \\
        \hline
        \multirow{4}{4em}{Multil. \mbox{S-BERT}} & \multirow{4}{1.5em}{0.27} & \textit{308} & \textit{\mean{-0.80}} & \textit{\contrib{0.01}} \\
        & & \textit{281} & \textit{\mean{0.67}} & \textit{\contrib{0.003}} \\
        & & \textit{176} & \textit{\mean{0.57}} & \textit{\contrib{0.002}} \\
        & & \textit{152} & \textit{\mean{-0.57}} & \textit{\contrib{0.002}} \\
    \end{tabular}
    \caption{Outliers and anisotropy scores in the output layer of each model. The numbers in this table are based on the Wikipedia data. Outliers are sorted by magnitude. We use the $3\sigma$ definition of outlier dimensions. We \textit{italicise} dimensions that do not qualify as outliers under the $5\sigma$ definition.
    }
    \label{tab:outlier-anisotropy-wiki}
\end{table*}

\begin{table*}[th]
    \centering
    \begin{tabular}{c|cccccc}
        \textbf{Model} & \textbf{ar} & \textbf{en} & \textbf{es} & \textbf{su} & \textbf{sw} & \textbf{tr} \\
        \hline
        XLM-R & 0.996 & 0.997 & 0.996 & 0.996 & 0.995 & 0.996 \\
        mBERT (cased) & 0.65 & 0.49 & 0.56 & 0.64 & 0.69 & 0.6 \\
        Multil. S-BERT & 0.21 & 0.17 & 0.19 & 0.28 & 0.59 & 0.17  \\
    \end{tabular}
    \caption{Anisotropy scores, final layer, per language, on the Wikipedia data.
    }
    \label{tab:anisotropy-wiki-per-lang}
\end{table*}

Results of the analysis are shown in Table~\ref{tab:outlier-anisotropy-1}.
XLM-R has an extremely high anisotropy score: Any given random sentence pair is already considered very similar to each other.
One of its outlier dimensions (588) contributes far and away the largest part to the expected cosine similarity.
This dimension is still present as an outlier, though with a smaller magnitude and cosine contribution, in the multilingual S-BERT which was derived from XLM-R.
The S-BERT model also has much lower anisotropy overall.

mBERT shows lower anisotropy than \mbox{XLM-R} but much higher values than the S-BERT.
Its two largest dimensions both contribute significantly to anisotropy.
Unlike \citet{rajaee-pilehvar-2022-isotropy}, we do find outlier dimensions in multilingual BERT.
It is worth noting that we use a different checkpoint than they do (they use the uncased model, we use the cased version), and we focus on sentence representations rather than individual word embeddings.
To verify our findings, we repeat our experiments on the same Wikipedia data they used---this now concerns the final layer of the model.
We calculate sentence embeddings in this case as well.
These results are listed in
Table~\ref{tab:outlier-anisotropy-wiki}.
Note that outlier dimensions can and do differ from layer to layer, which we observe in all three of these models.
The multilingual S-BERT has no outliers larger than $5\sigma$ in the output layer, but does have larger outlier dimensions in the middle layer 8.
It may be that the sentence-transformer tuning affects the later layers first and therefore more thoroughly.

In Table~\ref{tab:anisotropy-wiki-per-lang}, we report anisotropy scores per language for our models. 
We also use Wikipedia data here, since this includes fewer languages but is of a more natural domain than Tatoeba.
XLM-R exhibits such high anisotropy in these sentence embeddings that there is no meaningful difference between the scores across languages.
However, the other two models both show an interesting pattern:
English and Spanish have the most isotropic spaces, with anisotropy increasing roughly as training data size decreases.
This observation fits with the idea that anisotropy is frequency-driven \citep{yu-etal-2022-rare,puccetti-etal-2022-outlier}, i.e., that less frequent tokens tend to be pushed further from the origin.
Arabic is more anisotropic than Turkish despite having the same (S-BERT) or double (mBERT) the pre-training data size.
Presumably this is due to Arabic using a non-Latin script, since the model has seen more Latin-script data.
Sundanese and Swahili are the two languages with the smallest pre-training data of this set.
Swahili has the highest anisotropy in both models, and by a large margin in the S-BERT model.
This is somewhat surprising, since Sundanese has even smaller pre-training data, but may be down to data quality or tokenisation issues.
It may even be that the S-BERT tuning included bad Swahili data---however, this is speculation, since the relevant documentation is lacking.

For XLM-R, we further graph the average hidden representations per layer using Tatoeba data.
Layer 8 is shown in Figure~\ref{fig:tatoeba-mean-layer8}; all layers in Figure~\ref{fig:xlmr-tatoeba-means-all} in the Appendix.

\begin{figure}
    \centering
    \includegraphics[width=0.48\textwidth]{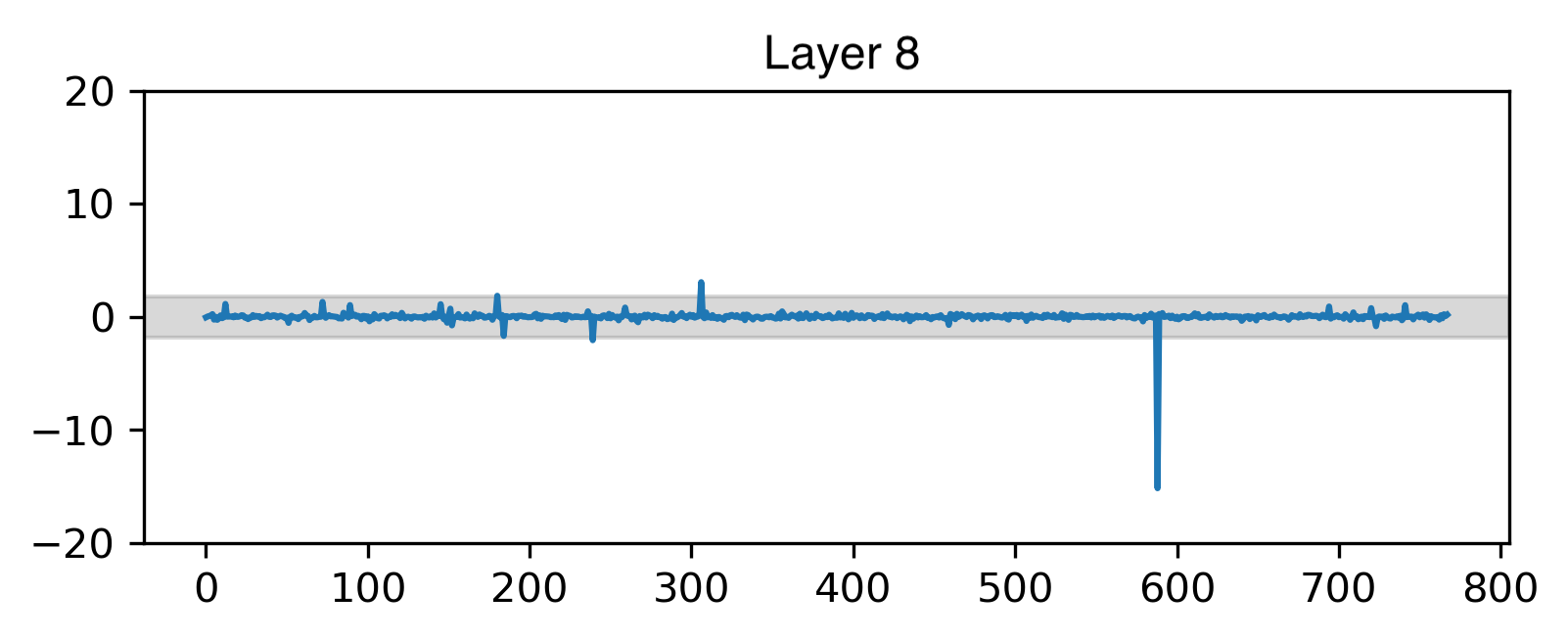}
    \caption{XLM-R mean embedding over all Tatoeba data; layer 8. The grey area denotes $3\sigma$ around the mean. The outlier dimensions are clearly visible.}
    \label{fig:tatoeba-mean-layer8}
\end{figure}

\section{Zeroing out dimensions}\label{sec:zeroing-dims}

Based on the outlier analysis, we experiment with zeroing out dimensions from the sentence representations before feeding them to the similarity search functions.
The biggest outlier, 588, clearly damages performance by greatly raising 
the
similarity of all sentences.
The correct candidate may thus be eclipsed by a false one more easily.
Figure~\ref{fig:cosines-588-306} illustrates how this 
occurs.
On the x-axis are the ranking positions of candidate sentences, on the y-axis their average cosine distances (inverse to cosine similarity).
In the unmodified model, all candidates are highly similar to the query sentence.
After removing 588, candidates with lower ranking become much more dissimilar, and the difference between the top candidate and the other sentences increases, which is a desirable property (note that 
the graphic does not show whether and which candidate sentences changed their ranking as a result).

In addition to zeroing the largest outliers, we identified other dimensions of interest by their magnitude.
We included the ten largest dimensions in each language of Tatoeba, finding a total of 18 dimensions that are in the top ten for any of the 36 languages.\footnote{[12, 63, 145, 151, 152, 266, 267, 459, 723, 728, 588, 306, 239, 184, 180]}
These dimensions include the outliers previously identified, as well as additional large dimensions.
We explored removing these dimensions individually and generally found smaller effects, though still a marked 
effect
for some of them.
The results are listed in Table~\ref{tab:tatoeba-outliers-removed}.

We removed the same dimensions from sentence embeddings of BUCC \citep{ZWEIGENBAUM18.12} data.
Interestingly, this sometimes improved precision while also worsening recall.
Thus, the overall improvements on this task were small (e.g., 588) or even negated (306).
Removing all 18 large dimensions from Tatoeba and BUCC 
yields +9.7 accuracy and +5.3 F1
over the vanilla XLM-R model, respectively.
That said, even with this performance gain, the gap to the sentence-transformer is still very large.
In addition, manually zeroing a large number of dimensions depending on the task data cannot be done in a real-world system. 

\begin{table}[t]
    \centering
    \begin{tabular}{c|cc}
        \textbf{Model} & \textbf{Tatoeba}  & \textbf{BUCC} \\
        \hline
         XLM-R               & 50.35 & 59.1 \\
         XLM-R \textit{-588} & 52.99 & 59.6 \\
         XLM-R \textit{-306} & 50.59 & 58.0 \\
         XLM-R \textit{-239} & 51.11 & 59.2 \\
         XLM-R, \textit{18 dims rem.} & \textbf{60.09} & \textbf{64.4} \\
         \hdashline
         Multil. S-BERT & 85.17 & 85.7 \\
    \end{tabular}
    \caption{Average Tatoeba (accuracy)
    and BUCC (F1) 
    scores for XLM-R and modified versions with large dimensions set to zero. The multilingual S-BERT is included as a reference.}
    \label{tab:tatoeba-outliers-removed}
\end{table}

\section{Isotropy-enhancing operations}\label{subsec:isotropy-enhancing}

\begin{table*}[th]
    \centering
    \begin{tabular}{c|c|c|c|c|c|c}
        \textbf{Model} & \textbf{Anisotropy} & \textbf{Tatoeba} & \multicolumn{4}{|c}{\textbf{STS}} \\
         & & & \textbf{ar-en} & \textbf{es-en a)} & \textbf{es-en b)} & \textbf{tr-en} \\
        \hline
         XLM-R & 0.92 & 50.35 & .114 & .04 & -.059 & .141 \\
         XLM-R, \textit{18 dims rem.} & 0.47 & 60.09 & --- & --- & --- & --- \\
         XLM-R + CBIE & $-3.9 \times 10^{-5}$ & 69.01 & .316 & \textbf{.445} & .121 & \textbf{.37} \\
         XLM-R + Whitening & $7.6 \times 10^{-5}$ & \textbf{70.03} & \textbf{.355} & .444 & \textbf{.153} & .36 \\
         \hdashline
         mBERT & 0.73 & 37.53 & .20 & .244 & .146 & .172 \\
         mBERT + CBIE & $5.7 \times 10^{-5}$ & \textbf{45.79} & \textbf{.25} & \textbf{.403} & .15 & \textbf{.217} \\
         mBERT + Whitening & $-6.6 \times 10^{-6}$ & 45.14 & .208 & .395 & \textbf{.171} & .154 \\
         \hdashline
         Multil. S-BERT & 0.35 & 85.17 & \textbf{.772} & \textbf{.779} & \textbf{.235} & \textbf{.762} \\
         S-BERT + CBIE & $5.8 \times 10^{-5}$ & 86.36 & .722 & .742 & .233 & .724 \\
         S-BERT + Whitening & 0.0001 & \textbf{87.35} & .745 & .772 & .222 & .748 \\
    \end{tabular}
    \caption{Anisotropy scores, average Tatoeba (accuracy) scores, and STS cross-lingual subset scores (Pearson correlation) for XLM-R, mBERT, multilingual S-BERT, and modified versions with post-hoc transformations applied to the sentence embeddings.}
    \label{tab:tatoeba-sts-transformed}
\end{table*}

Aside from directly zeroing out individual dimensions, we can apply transformations over the set of embeddings that largely eliminate anisotropy and mean-center the representations.
In this work, we test two such transformations:

\begin{enumerate}
    \item ZCA Whitening \citep[cf.][]{huang-etal-2021-whiteningbert-easy}
    \item Cluster-based isotropy enhancement \citep{rajaee-pilehvar-2021-cluster}
\end{enumerate}

\subsection{ZCA Whitening}

Whitening is an operation originally used in data pre-processing, in order to remove correlations between the input data features to a machine learning system.
It is also called a ``sphering transformation'', since the resulting data space is a hyperdimensional sphere.
However, whitening has recently been used to transform output embeddings of models such as BERT \citep[cf.][]{huang-etal-2021-whiteningbert-easy}, before using them for downstream applications.

For a given space $X$ with covariance $\Sigma$ and mean 0,
there are many 
valid whitening transformations.
The resulting matrix $Y = WX$ must have the identity matrix $I$ as its covariance, and the whitening transformation $W$ must satisfy the condition:

\begin{equation}
     W^{T} W = \Sigma^{-1}.
\end{equation}

Given that $\Sigma$ can be decomposed into:
\begin{equation}
    \Sigma = D\Lambda D^{T},
\end{equation}
a valid $W$ can be found as follows:
\begin{equation}
    W = D\Lambda^{-\frac{1}{2}}D^{T}.
\end{equation}

\subsection{Cluster-based isotropy enhancement}

We adopt this method from \citet{rajaee-pilehvar-2021-cluster}.
The first step is to separate the provided data into clusters. 
In their paper, \citet{rajaee-pilehvar-2021-cluster} use 27 clusters.
We make the number of clusters dependent on the number of examples---with too few examples in a single cluster, the concept of ``isotropy'' becomes meaningless, and it can lead to computation errors.
Each cluster is mean-centered, which is necessary for the subsequent steps.  
Then, PCA is applied to every cluster, and the top k principal components (``dominant directions'') are zeroed out. 
We follow the original paper in setting $k=12$.

\subsection{Discussion}

The common thread of these methods is that they transform the output representations based on some set of encoded data.
This means that either the transformation must be calculated anew for every set of data, or retained from a training set in order to apply it to new data.
Though this is not ideal from an application perspective, we follow the approach of calculating the transformation for every new set of encoded data.
The tasks in question do not use fine-tuning on any kind of training data, so we transform the embedded test data.
An alternative would be to learn and retain a transformation based on some external dataset, then apply this to the task data.
Such an approach would be especially helpful when doing inference on only a few queries at a time, or when the overhead of computing the transformation should be avoided at inference time.

\subsection{Results}

After applying the transformations, we run our anisotropy analysis again.
We also test Tatoeba and STS performance before and after the transformations.
The results are listed in Table~\ref{tab:tatoeba-sts-transformed}.
For XLM-R, the transformations lead to a performance boost of almost 20 points on Tatoeba.
Recall that removing the top dimensions improved accuracy by only around 10 points.
For mBERT, which is more isotropic to begin with, the difference is only eight points.
Other factors, such as a more complex misalignment of different languages, seem to be a bigger bottleneck for its performance.
The multilingual S-BERT benefits very little from the isotropy-enhancing transformations.

For STS, the multilingual S-BERT in fact performs better without the transformations.
mBERT and XLM-R do benefit from the transformations to some degree: In most cases, there is a large improvement, particularly in XLM-R.
For mBERT, the \textbf{es-en b)} subset only shows a small improvement, and the others benefit more from CBIE than from whitening.
\citet{rajaee-pilehvar-2022-isotropy} also test on STS, including the monolingual subsets.
However, since they report Spearman correlations rather than Pearson, as well as using a different mBERT checkpoint than we do, the numbers are not directly comparable, and we do not show them in our table.
The main takeaway here is that using the whitening transformation yields similar results overall to CBIE, and that both work to improve sentence-level representations for semantic similarity.
Also, they both have little to no benefit in the S-BERT model, which was tuned with parallel data and is already much more isotropic.

After the transformations, anisotropy scores are very close to zero; that is, the spaces are extremely isotropic.
We can also see this in the t-SNE visualisations of these spaces, see \S~\ref{subsec:tsne}.
However, applying the outlier definition of three times the standard deviation, we still find outlier dimensions in the transformed spaces.
These all have very small magnitude, and are not necessarily related to the dimensions that were outliers before.
Since the transformations are not deterministic, these outlier dimensions can also change when recalculating the transformed spaces.
Therefore, we do not consider these dimensions true outliers.
In an (artificially) highly isotropic space, the traditional outlier definition of larger than three standard deviations may simply not apply.

\section{Embedding space visualisation}\label{subsec:tsne}

\begin{figure*}
    \centering
    \hspace{10pt}
    \begin{minipage}{.32\textwidth}\centering Arabic-English\end{minipage}%
    \begin{minipage}{.32\textwidth}\centering Bengali-English\end{minipage}%
    \begin{minipage}{.32\textwidth}\centering German-English\end{minipage}%

    \rotatebox{90}{\begin{minipage}{102pt}\centering Unmodified\end{minipage}}%
    \includegraphics[width=0.32\textwidth]{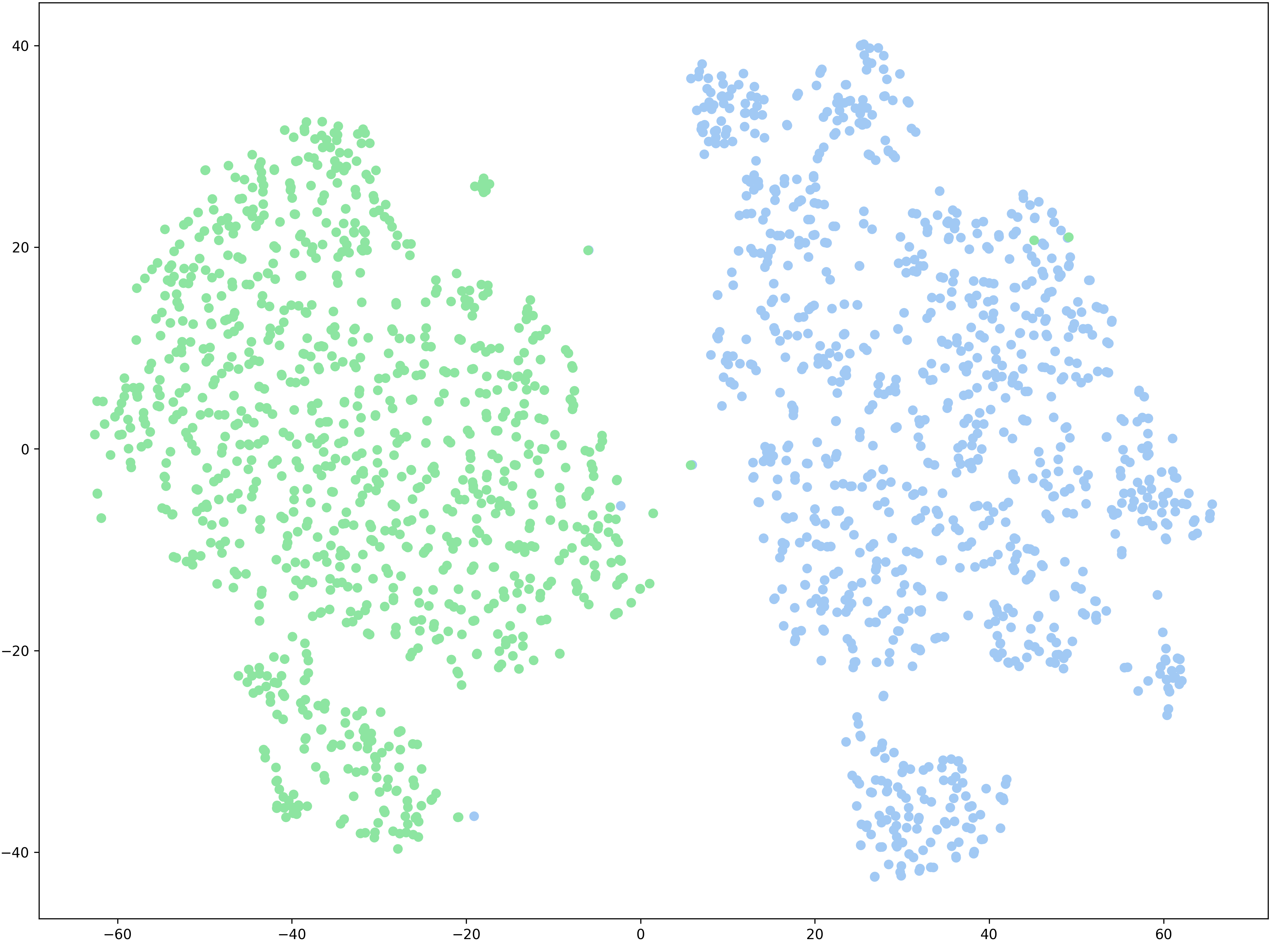}%
    \includegraphics[width=0.32\textwidth]{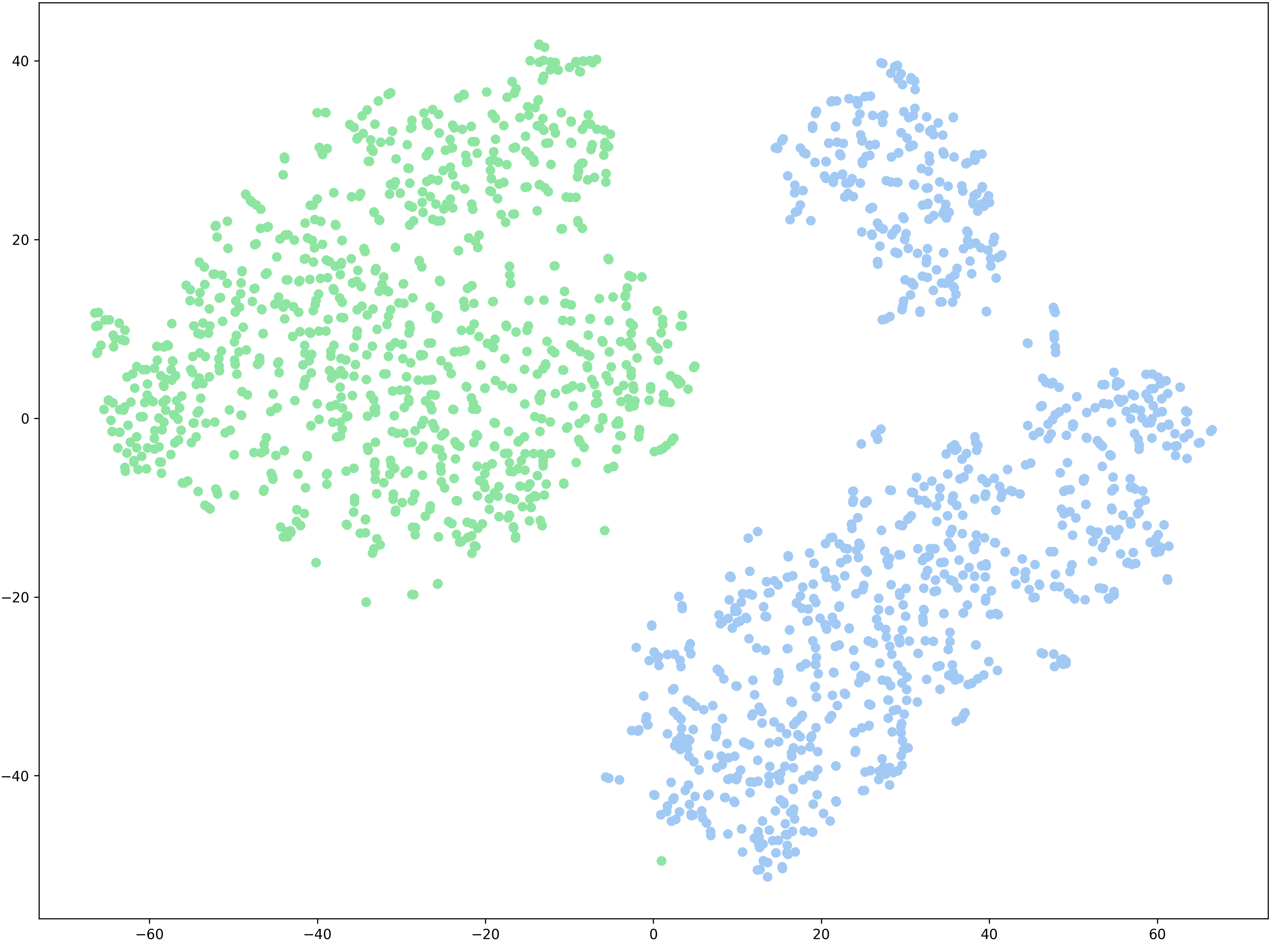}%
    \includegraphics[width=0.32\textwidth]{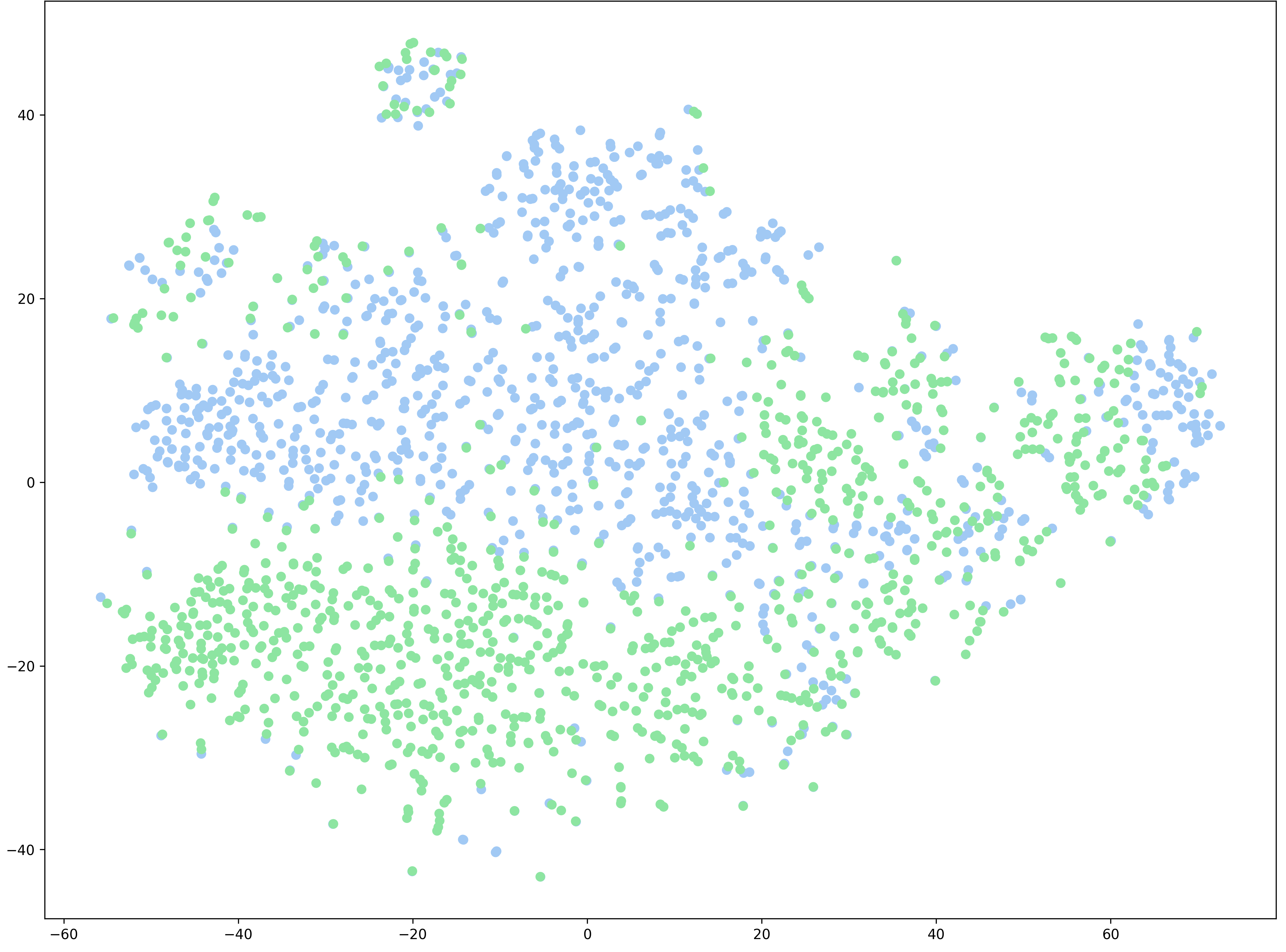}%
    
    \rotatebox{90}{\begin{minipage}{102pt}\centering CBIE-transformed\end{minipage}}%
    \includegraphics[width=0.32\textwidth]{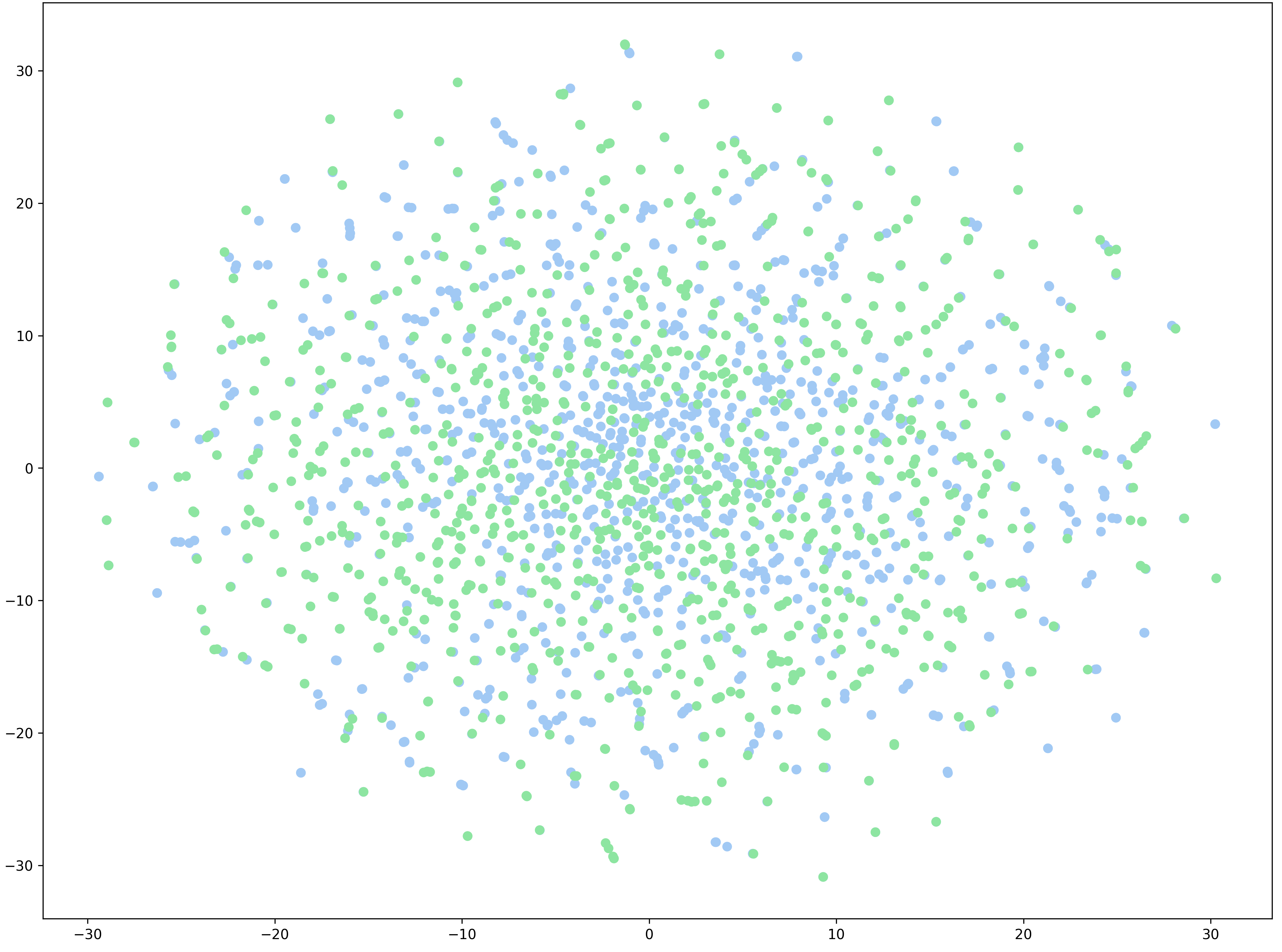}%
    \includegraphics[width=0.32\textwidth]{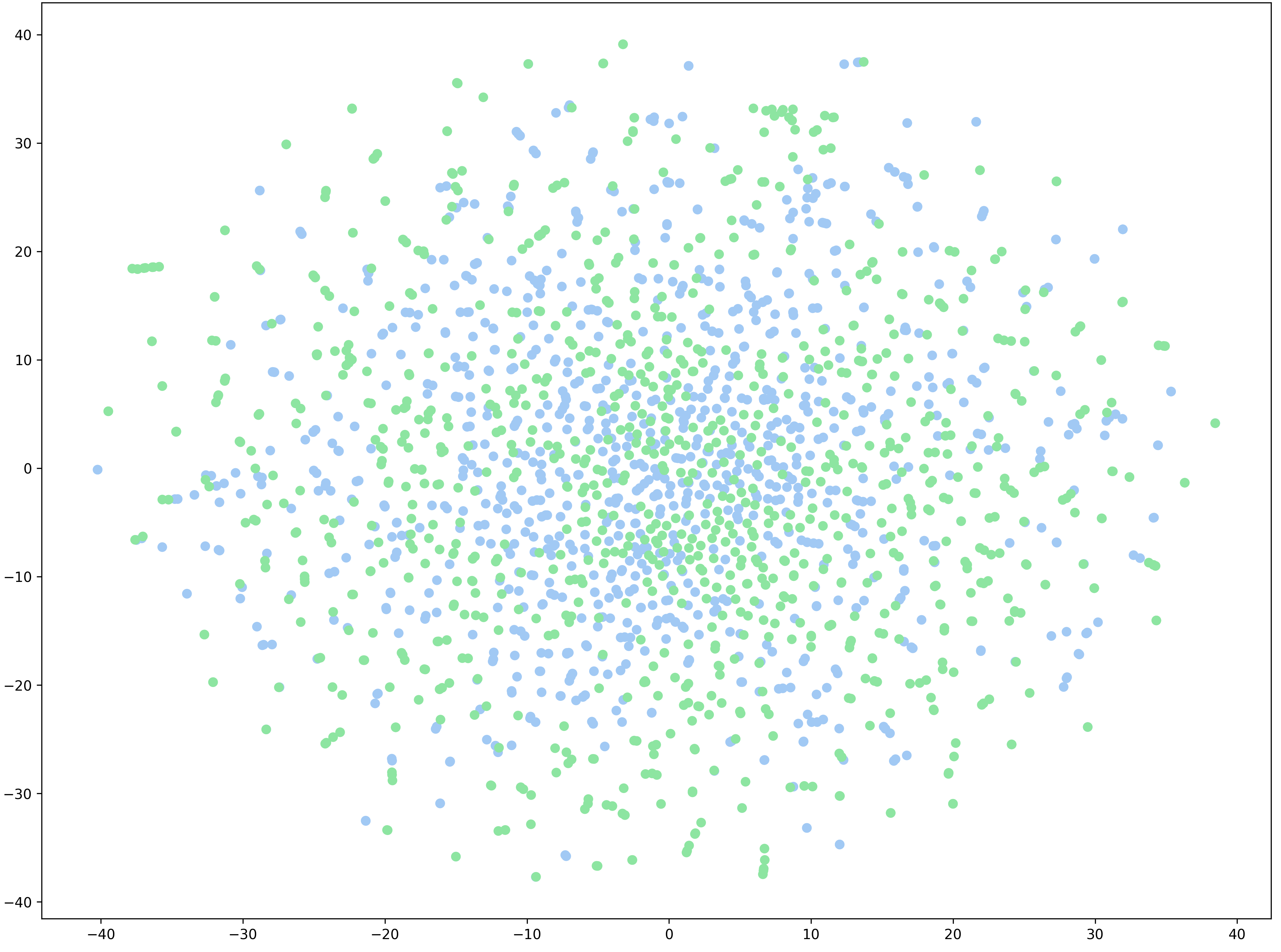}%
    \includegraphics[width=0.32\textwidth]{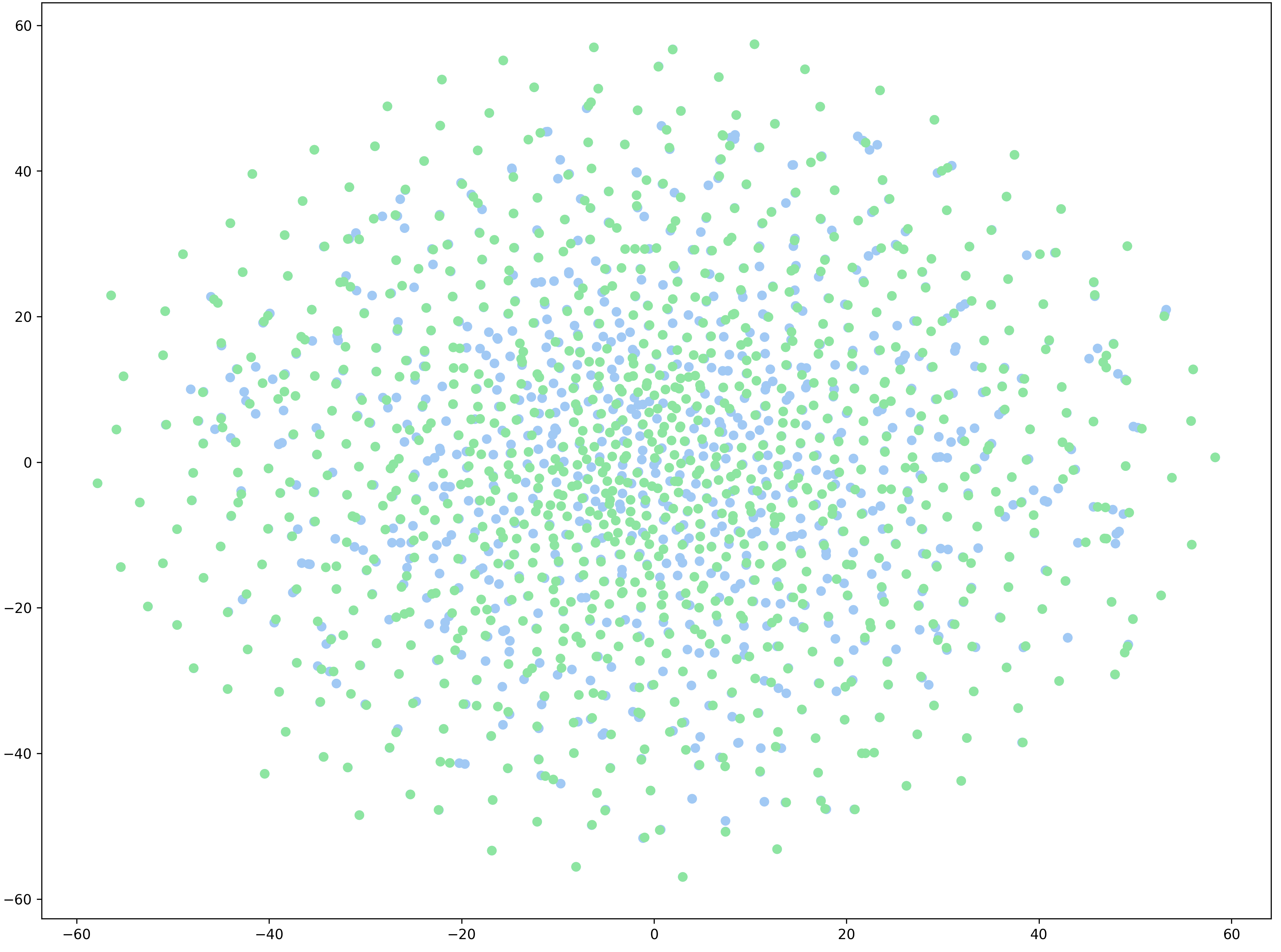}%

    \caption{Left to right: Arabic-English, Bengali-English, and German-English Tatoeba sentence embeddings from XLM-R. Top to bottom: Unmodified and CBIE-transformed
    versions of the embeddings.
    The source languages are shown in green, the English in blue.}
    \label{fig:tsnes}
\end{figure*}

To visualise the representation space, we use t-SNE \citep{Maaten2008VisualizingDU}.
First, we apply a PCA dimensionality reduction to 50 dimensions.
Then, we reduce the dimensionality further using t-SNE and plot the space in two dimensions.
In Figure~\ref{fig:tsnes}, we show examples from Tatoeba data in XLM-R: Arabic, Bengali, and German.
For the first two, accuracy increased by more than 20 points after the transformation, while German is already a high-resource language where accuracy only increased by around 5 points.
Since CBIE and Whitening produce very similar visualisations, we only show CBIE.

The unmodified spaces very clearly show the problem of internal misalignment between different languages in the model, which disproportionately affects languages with less pre-training data and/or non-Latin scripts.
With Arabic-English and Bengali-English, the source and target language spaces are almost disjunct.
This issue can be addressed using isotropy-increasing transformations, but they do not solve the problem entirely.
For instance, the unmodified sub-spaces of Bengali and English also have markedly different shapes, despite representing a set of parallel sentence pairs.
Matching the equivalent sentences to each other starting from such different spaces is more complex than merely applying a linear transformation to increase isotropy.

\section{Conclusions}

We have analysed how outlier dimensions and anisotropy interact with cross-lingual semantic similarity tasks in pre-trained multilingual language models.
In particular, we focused on the sentence representations of multilingual BERT and \mbox{XLM-R}, comparing them to the sentence representations of a multilingual S-BERT model---essentially a modified XLM-R trained with parallel data to optimise for sentence representations.
We employed a range of methods on several different tasks to approach the question from multiple angles.
The simplest method of increasing isotropy is removing the largest (outlier) dimensions from the sentence embeddings.
We compared the results of this with further-reaching isotropy-increasing transformations.
Additionally, we examined how changing the representations affected anisotropy measures and outlier dimensions.
Finally, we plotted unmodified and transformed sentence representation spaces to illustrate how anisotropy is one aspect that affects sentence similarity, but reducing it does not resolve all issues in the space.

\paragraph{Future Work.}

Potential future research questions include:
Are outliers and anisotropy also relevant when using \textit{fine-tuned} models for cross-lingual transfer? 
Do larger, particularly generative models, have these issues affecting cross-lingual similarity?
Are the pre-training dynamics of anisotropy in multilingual models similar to those of monolingual models?
How can we train multilingual models 
to avoid a degenerating representation space?

\section*{Limitations}

This paper examines the anisotropy and outlier phenomenon only for a few, relatively similar, models. 
The isotropy-increasing transformations are non-deterministic and have to be calculated post-hoc based on some set of embedded data, which may not be practical for
applications where inference is done on individual or small batches of examples.

Since we specifically consider sentence representations, we first average over word embeddings before calculating the mean and standard deviation for outlier analysis.
This in effect reduces the sample size and leads to a smaller standard deviation, making our analysis more sensitive to even slight outlier dimensions.
Another reason to work with relatively small datasets is to make computing the transformations simple and fast, but this may limit the ability of these transformations to generalise.

\section*{Acknowledgements}

This publication was supported by LMUexcellent, funded by the Federal Ministry of Education and Research (BMBF) and the Free State of Bavaria under the Excellence Strategy of the Federal Government and the Länder; and by the German Research Foundation (DFG; grant FR 2829/4-1).
The work at CUNI was supported by Charles University project PRIMUS/23/SCI/023, and by the European Commission via its Horizon research and innovation programme (No. 870930 and 101070350).

\bibliography{anthology,custom}
\bibliographystyle{acl_natbib}

\appendix

\section{XLM-R Mean Embeddings of Tatoeba in all Layers}\label{app:tatoeba-all-means}

\begin{figure*}
    \centering
    \includegraphics[width=0.49\textwidth]{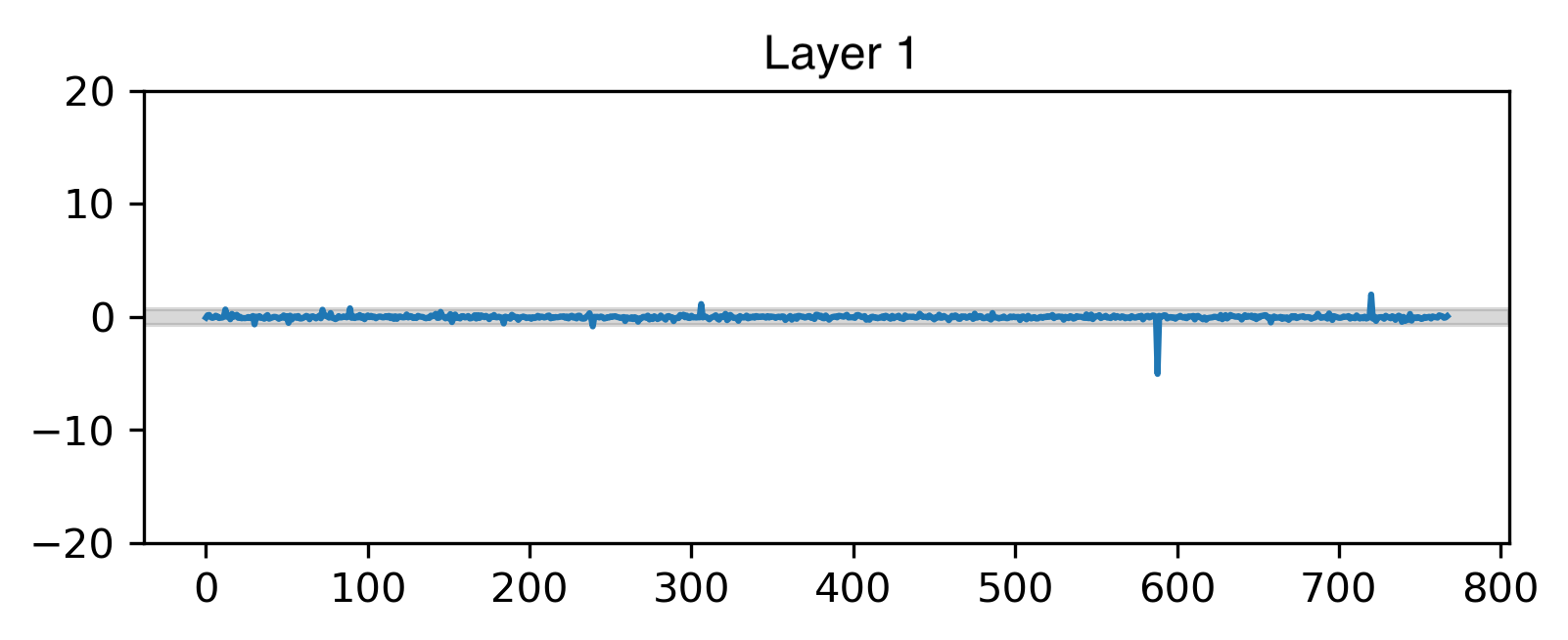}
    \includegraphics[width=0.49\textwidth]{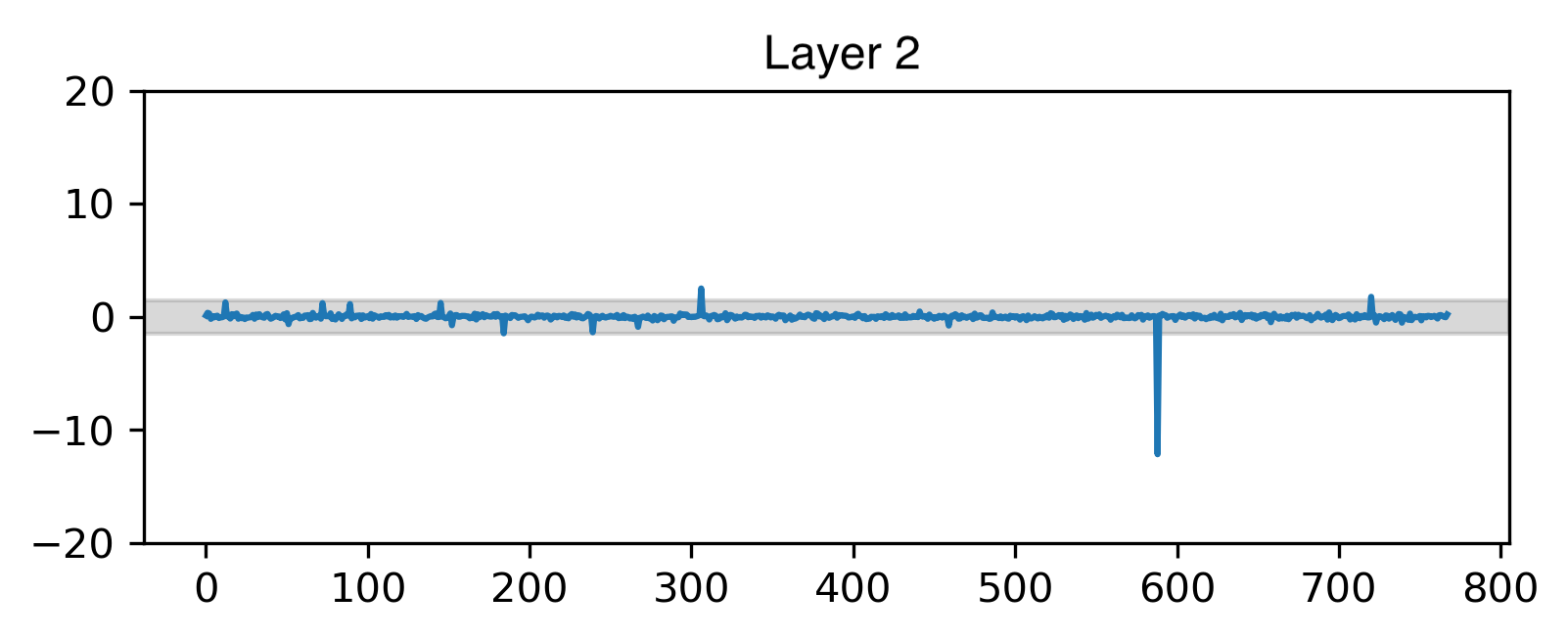}
    \includegraphics[width=0.49\textwidth]{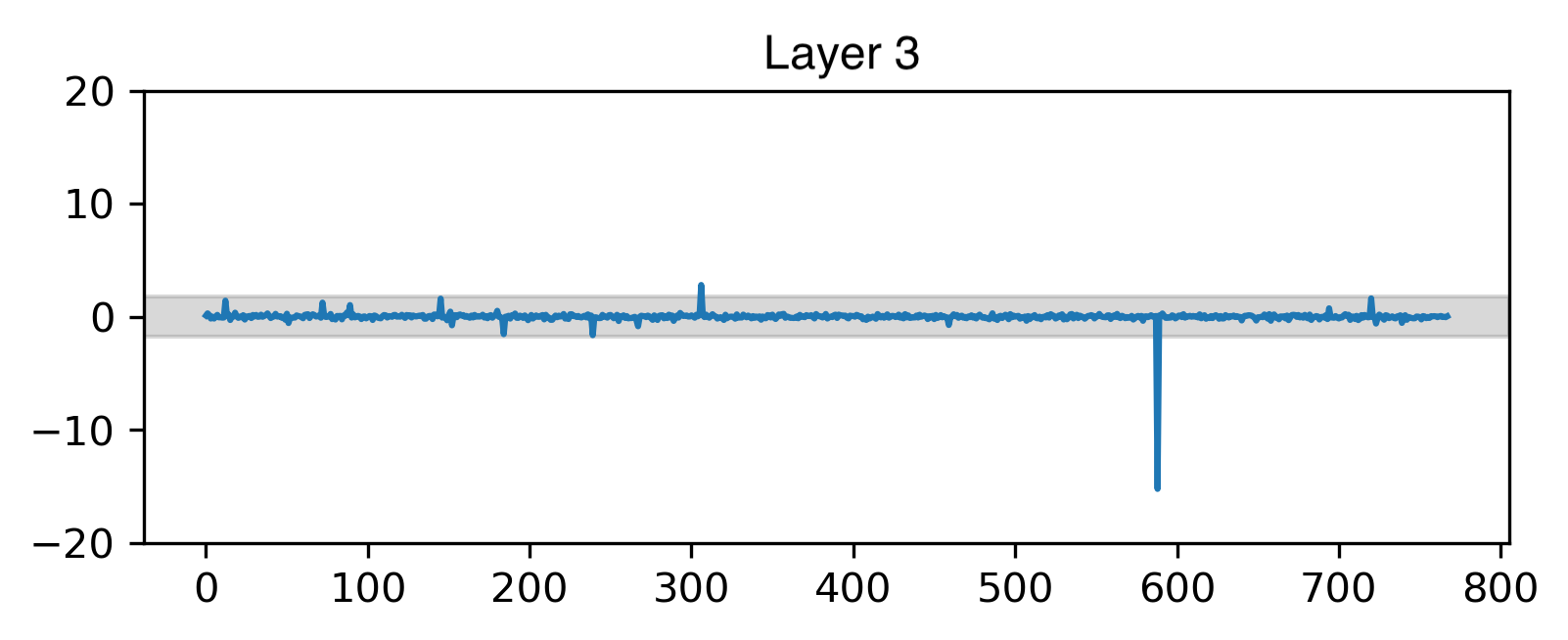}
    \includegraphics[width=0.49\textwidth]{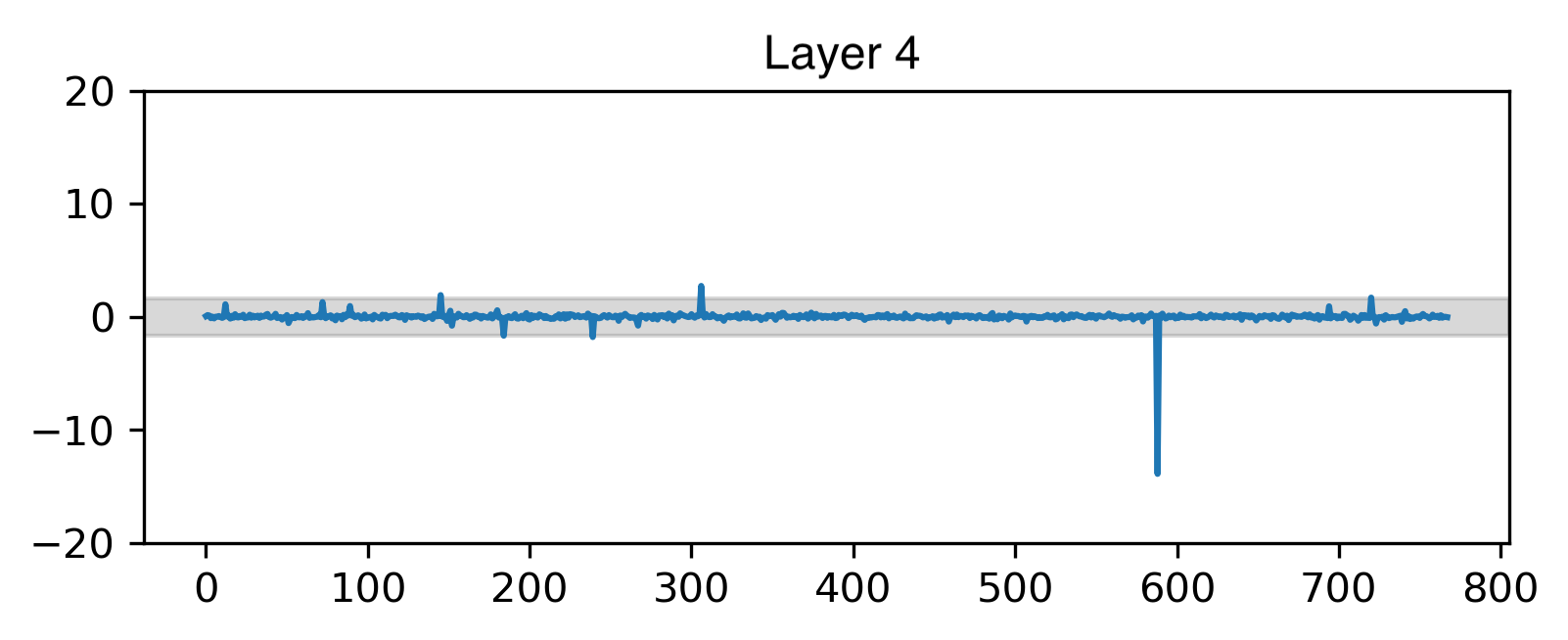}
    \includegraphics[width=0.49\textwidth]{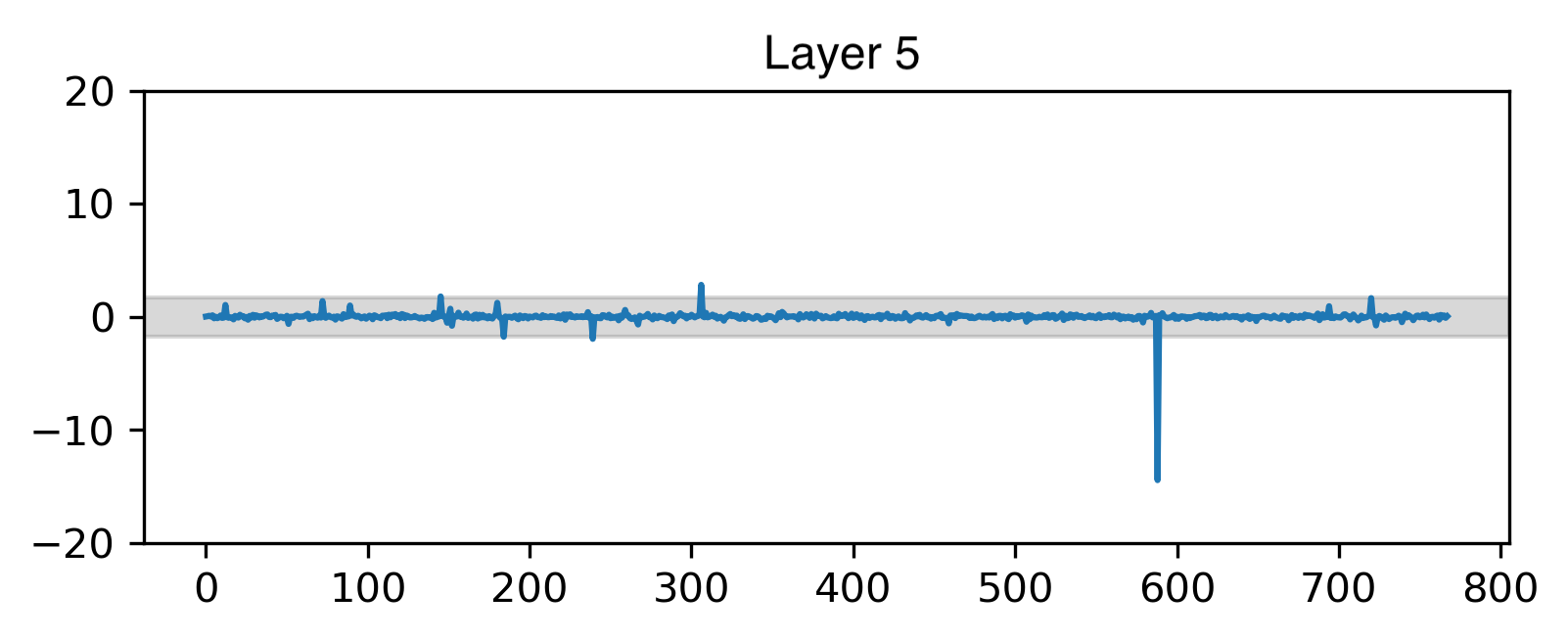}
    \includegraphics[width=0.49\textwidth]{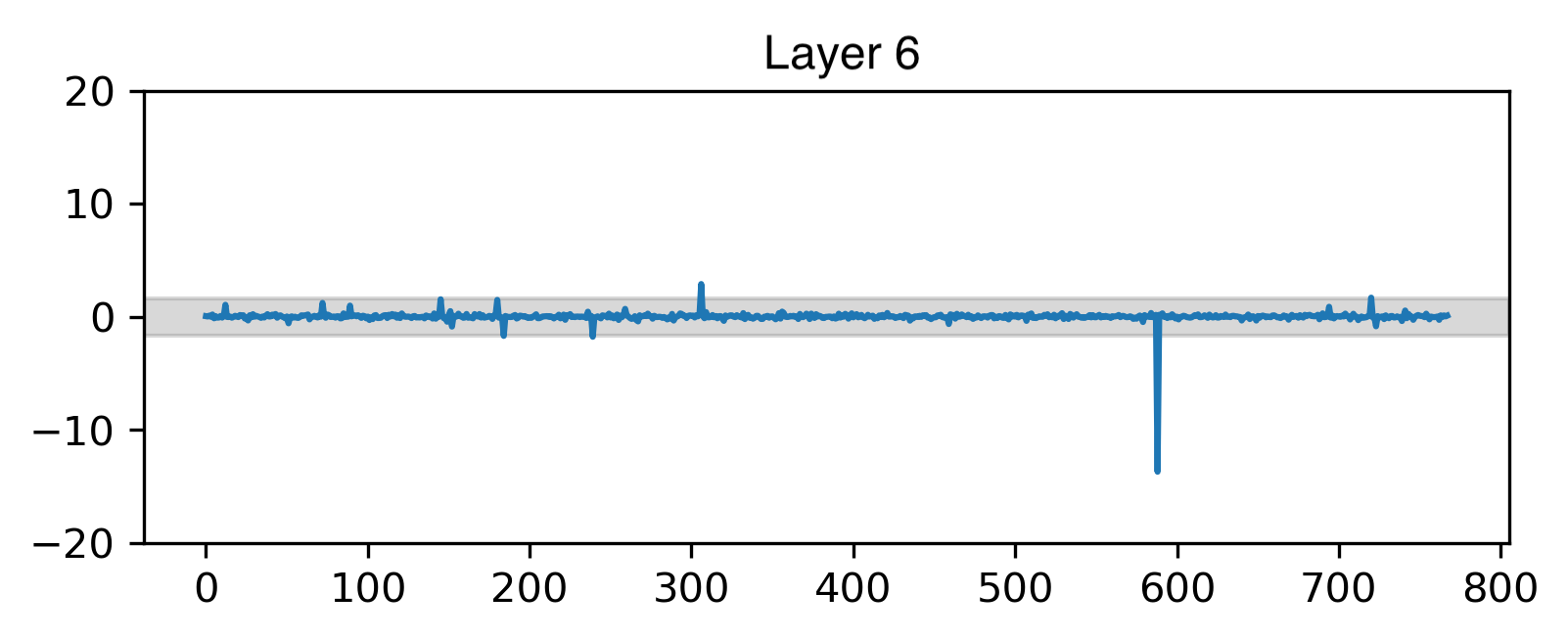}
    \includegraphics[width=0.49\textwidth]{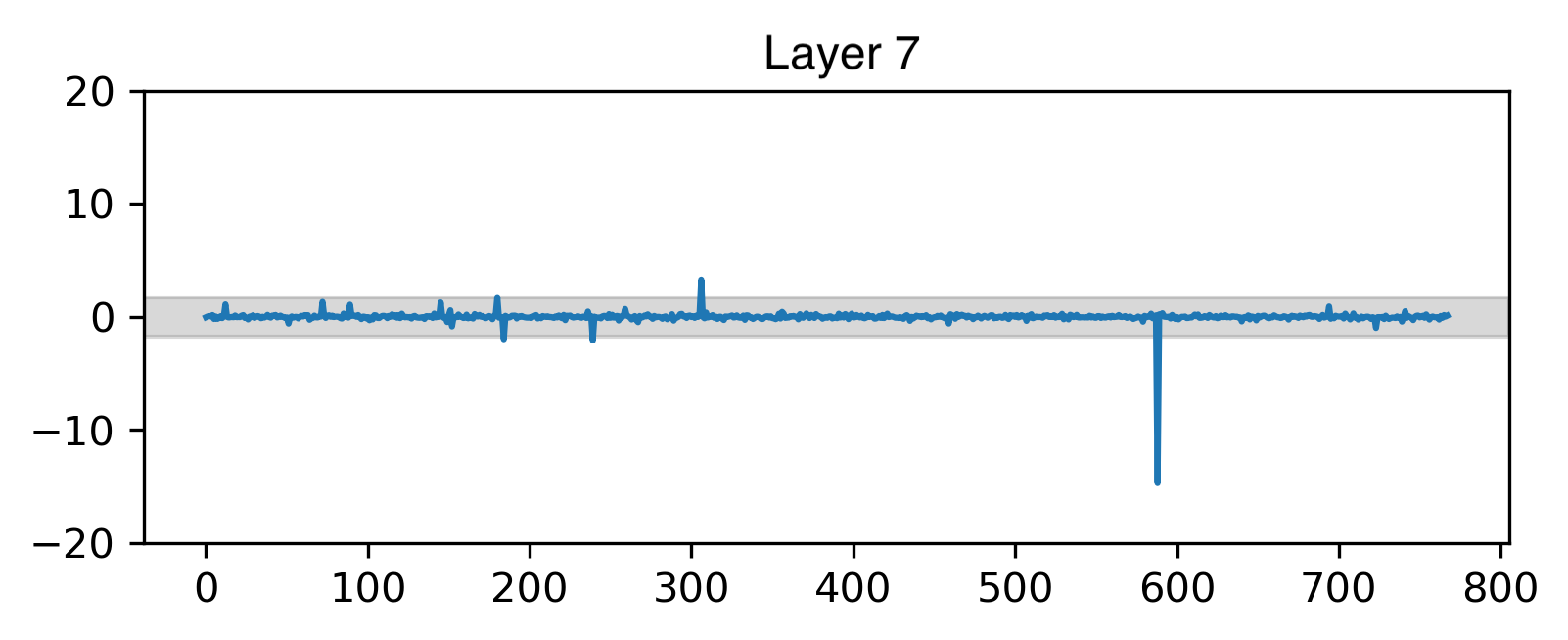}
    \includegraphics[width=0.49\textwidth]{img/xlmr_means_all_tatoeba/outliers_mean_layer_8.png}
    \includegraphics[width=0.49\textwidth]{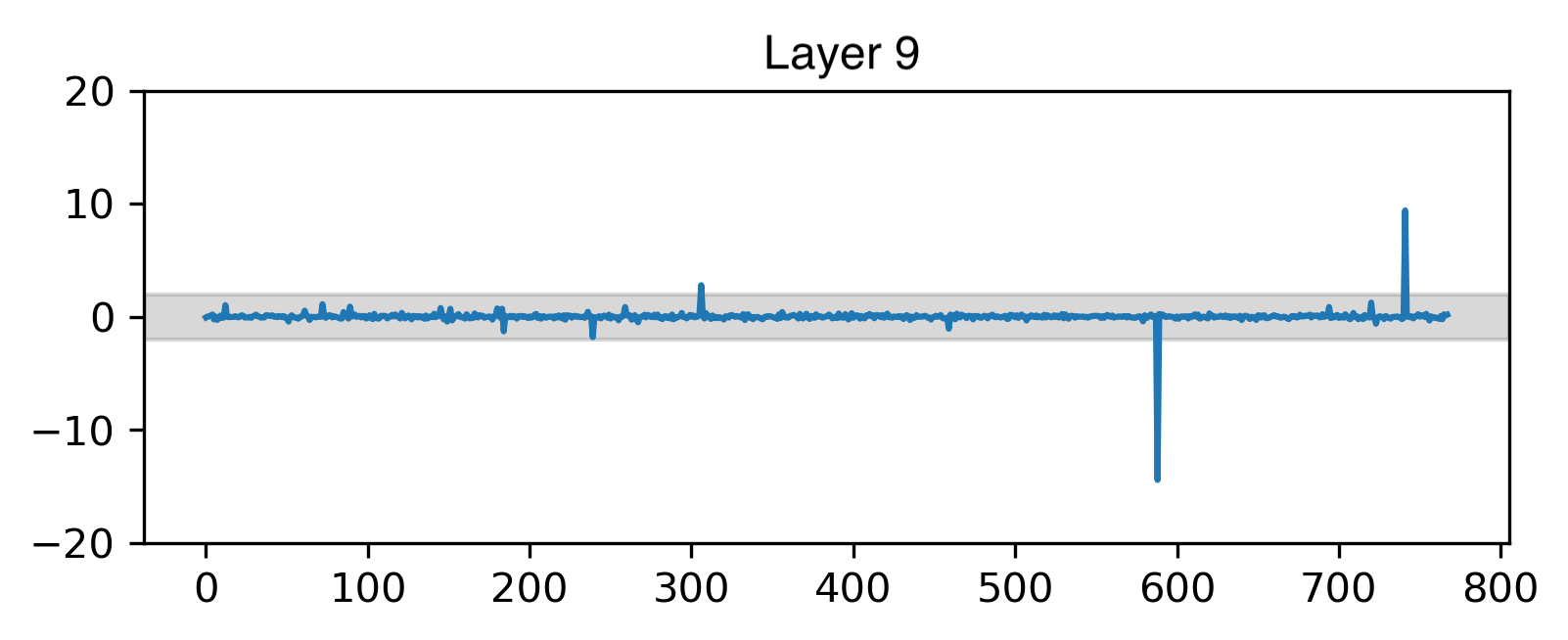}
    \includegraphics[width=0.49\textwidth]{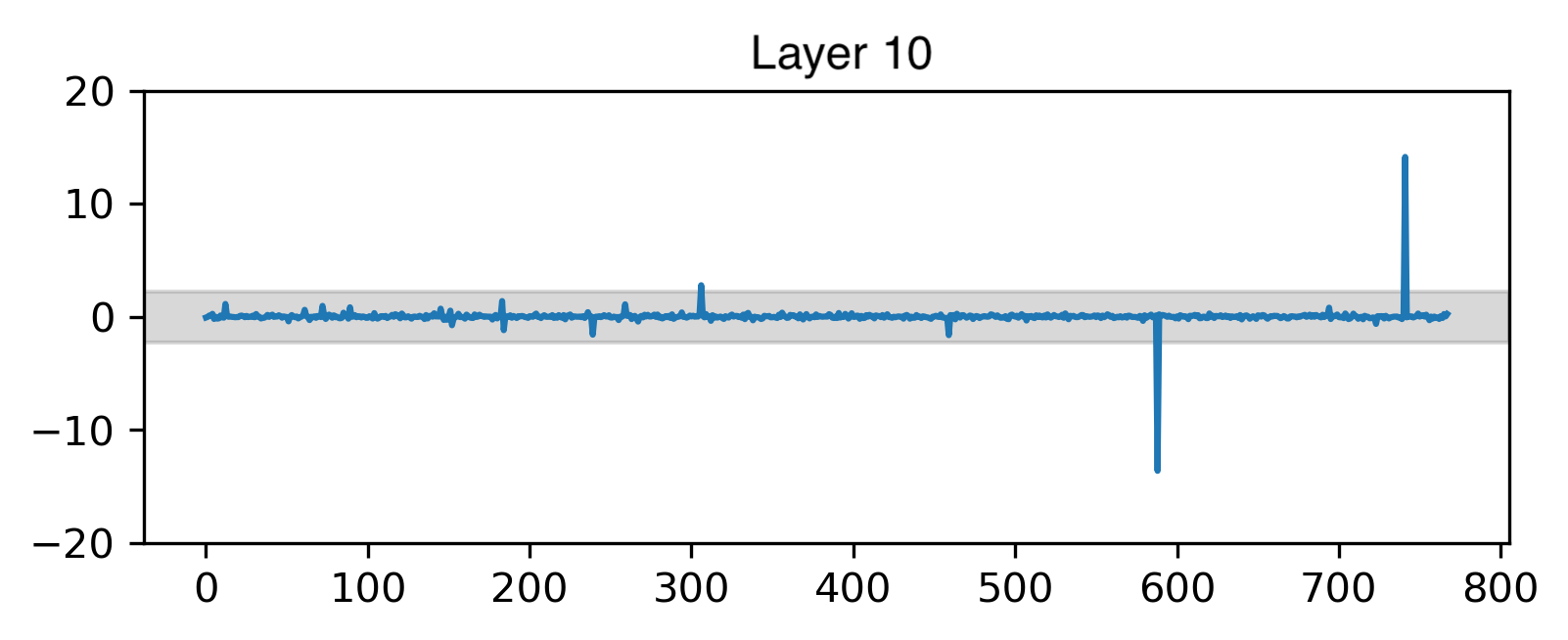}
    \includegraphics[width=0.49\textwidth]{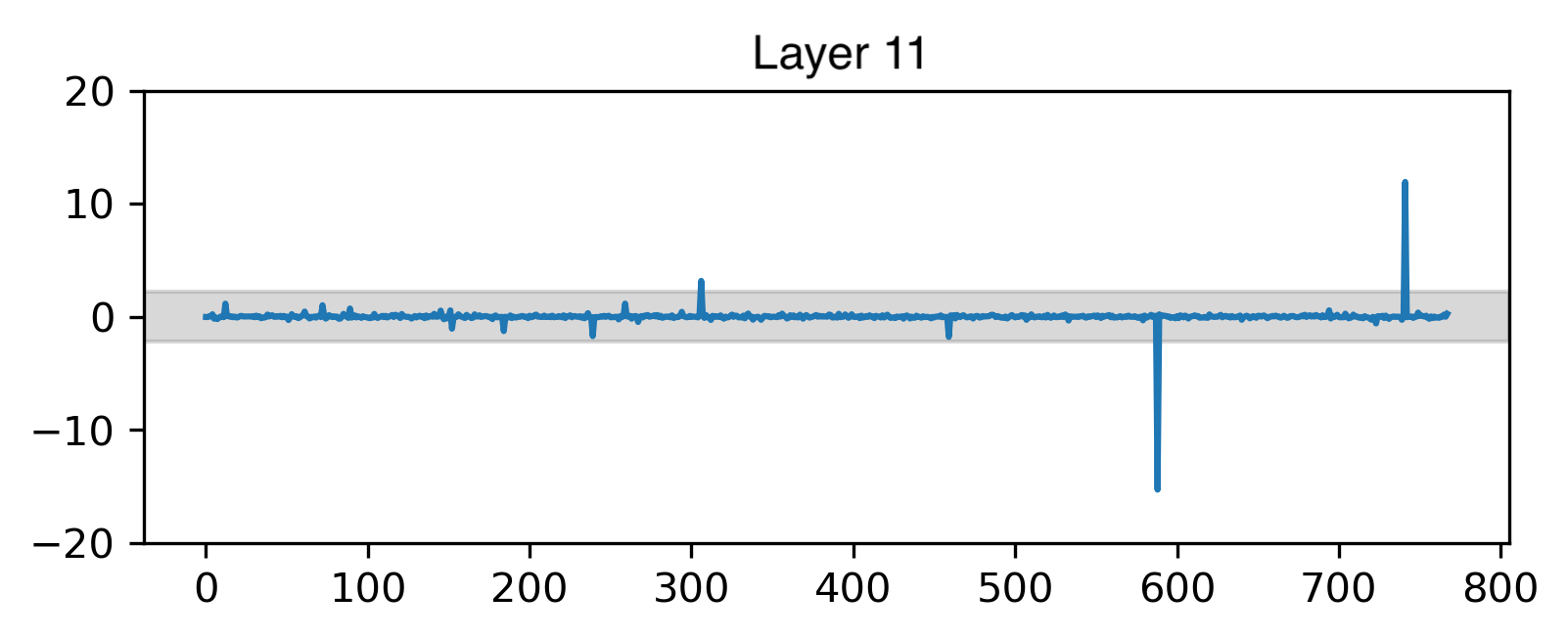}
    \includegraphics[width=0.49\textwidth]{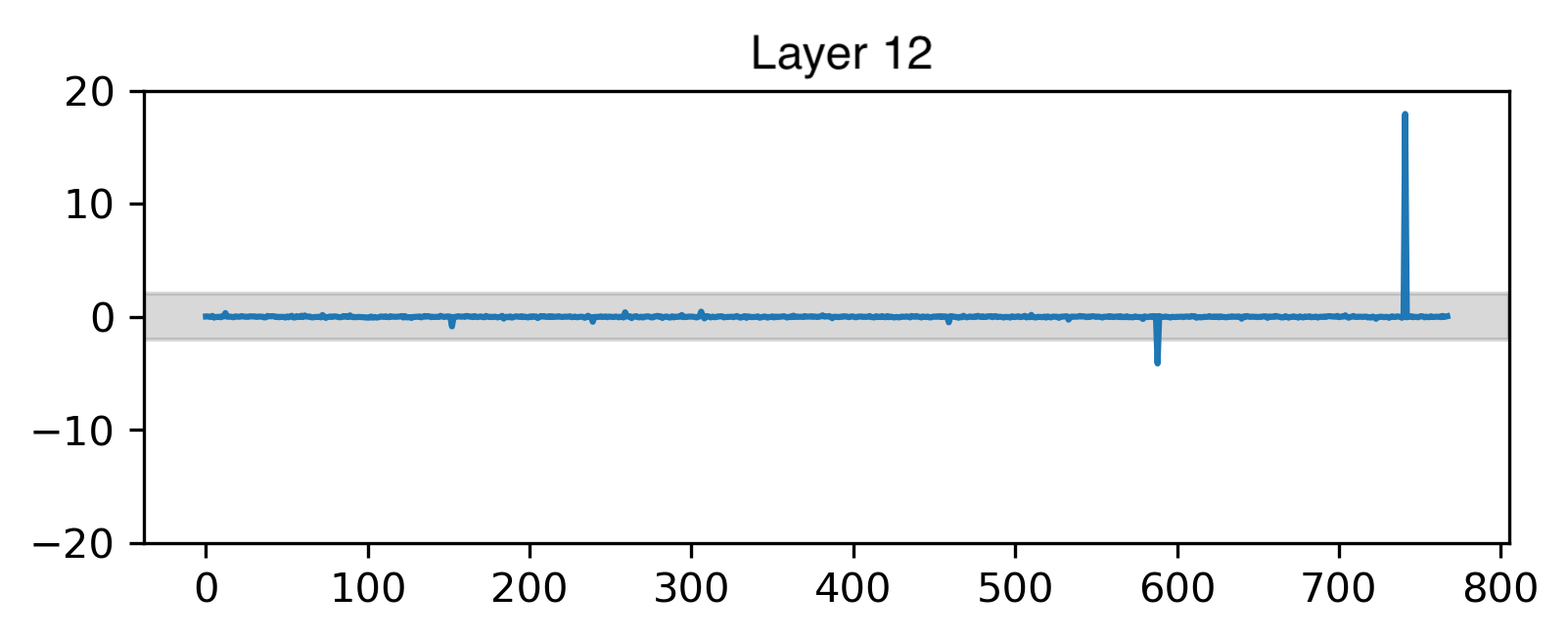}
    \caption{XLM-R mean embeddings on Tatoeba data.}
    \label{fig:xlmr-tatoeba-means-all}
\end{figure*}

See Figure~\ref{fig:xlmr-tatoeba-means-all}.

\end{document}